# Deep Learning Approach to Bearing and Induction Motor Fault Diagnosis via Data Fusion

Mert Sehri[1*], Merve Ertargin[2], Ozal Yildirim[2], Ahmet Orhan[2], Patrick Dumond[1]
[1]Department of Mechanical Engineering, University of Ottawa, Ottawa, Canada
[2]Department of Electrical Engineering, University of Firat, Elazig, Turkiye
*msehr006@uottawa.ca

***Abstract***—Convolutional Neural Networks (CNNs) are used to evaluate accelerometer and microphone data for bearing and induction motor diagnosis. A Long Short-Term Memory (LSTM) recurrent neural network is used to combine sensor information effectively, highlighting the benefits of data fusion. This approach encourages researchers to focus on multi model diagnosis for constant speed data collection by proposing a comprehensive way to use deep learning and sensor fusion and encourages data scientists to collect more multi-sensor data, including acoustic and accelerometer datasets.

## I. Introduction

Deep learning is a field of machine learning with applications such as machine diagnosis. There is an increased interest in collecting large datasets for bearing and induction motor fault diagnosis. However, only a limited number of vibration and acoustic datasets are publicly available for constant-speed data collection. This paper proposes a methodology to assess the effectiveness of accelerometer and microphone data using convolutional neural networks (CNNs) [1], [2], [3]. The utilization of CNNs in the methodology serves as a solution, allowing the model to extract intricate patterns and features from the available data, even in instances where the dataset is constrained. Unlike traditional methods that may struggle with small datasets, CNNs excel in learning hierarchical representations, enabling them to generalize well and make informed predictions despite limited samples. The datasets used to verify this approach are listed in Table I. The UORED-VAFCLS dataset consists of healthy, developing faults, and faulty bearings. A total of 20 bearings were tested and for each fault condition there are 5 bearings data with a sampling rate of 42 kHz and 420,000 samples focused on mechanical signals while the dataset consists of constant speed [4]. Additionally, UOEMD-VAFCVS dataset consist of induction motor data. There were 7 different electric motor faults and 1 healthy motor. The induction motor dataset has a sampling rate collected at 42 kHz and 420,000 samples which consists of electrical signals and includes constant speed and variable speed data [5]. Both the datasets consists of signal from vibration, acoustic, and temperature sensors. The vibration and acoustic sensors will be used to assess the effectivenes of the methodology. The proposed methodology is also extended to demonstrate the effectiveness of combining data from two sensors of different types using data fusion [6]. In this case, a long short-term memory (LSTM) [7] approach is used, which is a type of recurrent neural network, to take both data streams into account. The LSTM are a type of recurrent neural network (RNN) architecture. While CNNs are primarily used for tasks related recognition, LSTMs are used for sequence data which make them more suitable for tasks like speech recognition, natural language processing, and time series analysis [8].

TABLE I. Multi Sensor Datasets

| **Dataset Name** | ***Signal type*** | ***Sampling Frequency (kHz)*** | ***# of samples*** |
|---|---|---|---|
| UORED-VAFCLS [4], [9] | Vibration, acoustic, temperature | 42.0 | 420,000 |
| UOEMD-VAFCVS [5], [10] | Vibration, acoustic, temperature | 42.0 | 420,000 |

## II. Methodology

### *A. Neural Network Architecture for Vibration Data*

The proposed neural network architecture for the vibration data used in this study is shown in Fig. 1; the algorithm is a custom CNN designed for multi-class classification tasks with nine classes. The model begins with a series of convolutional layers (Conv1D) interlinked with max-pooling layers (MaxPool) to extract hierarchical features from the input time domain data. The convolutional layers have varying filter sizes and depths, with rectified linear unit (ReLU) activation functions. The network then flattens the output, which is then fed into two dense layers, the first with 32 neurons and ReLU activation, and the second consisting of nine neurons with softmax activation, producing a probability distribution over the nine classes. The architecture aims to capture patterns in the input data through convolutional operations and learn features for accurate classification. The model is trained using categorical cross entropy as the loss function, and the summary provided in Fig. 1 includes details such as input shape of 1000 x 1, output shapes, parameters, and activation functions for each layer.

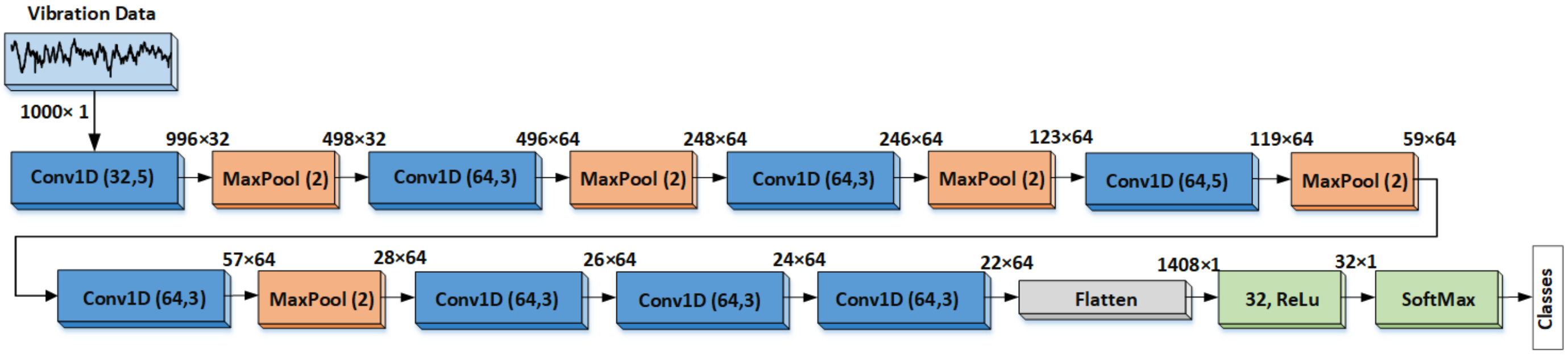

Figure 1. Neural Network Architecture for Vibration Data

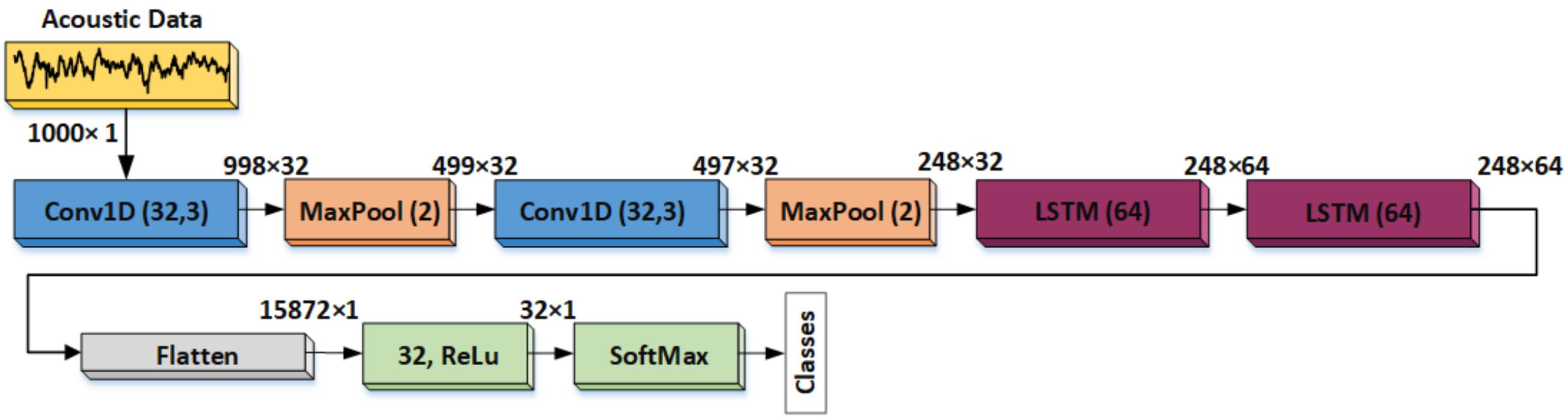

Figure 2. Neural Network Architecture for Vibration Data

### *B. Neural Network Architecture for Acoustic Data*

The model, shown in Fig. 2, is a custom CNN and RNN designed for multi-class classification of acoustic data with nine classes. It has Conv1D layers with varying filter sizes and depths, followed by maxpool layers for feature extraction. The network incorporates two LSTM layers with 64 filters, leading to a flatten layer. Two dense layers follow, with the first having 32 neurons and ReLU activation, and the second having a dense layer with nine neurons and softmax activation, yielding a probability distribution. The model in Fig. 2 shows the details of each layer's output shapes, parameters, and activation functions.

### *C. Neural Network Architecture for Combined Vibration and Acoustic Data*

Fig. 3 shows the proposed algorithm model that uses two parallel branches for processing acoustic and vibration data using CNN and RNN algorithms. This neural network architecture utilizes convolutional and recurrent networks to capture acoustic and vibration data properties. This architecture allows the training of the acoustic and vibration data inputs simultaneously using two different algorithms combined from Fig. 1 and Fig. 2 which will allow the model to capture different data types effectively and have a better classification results.

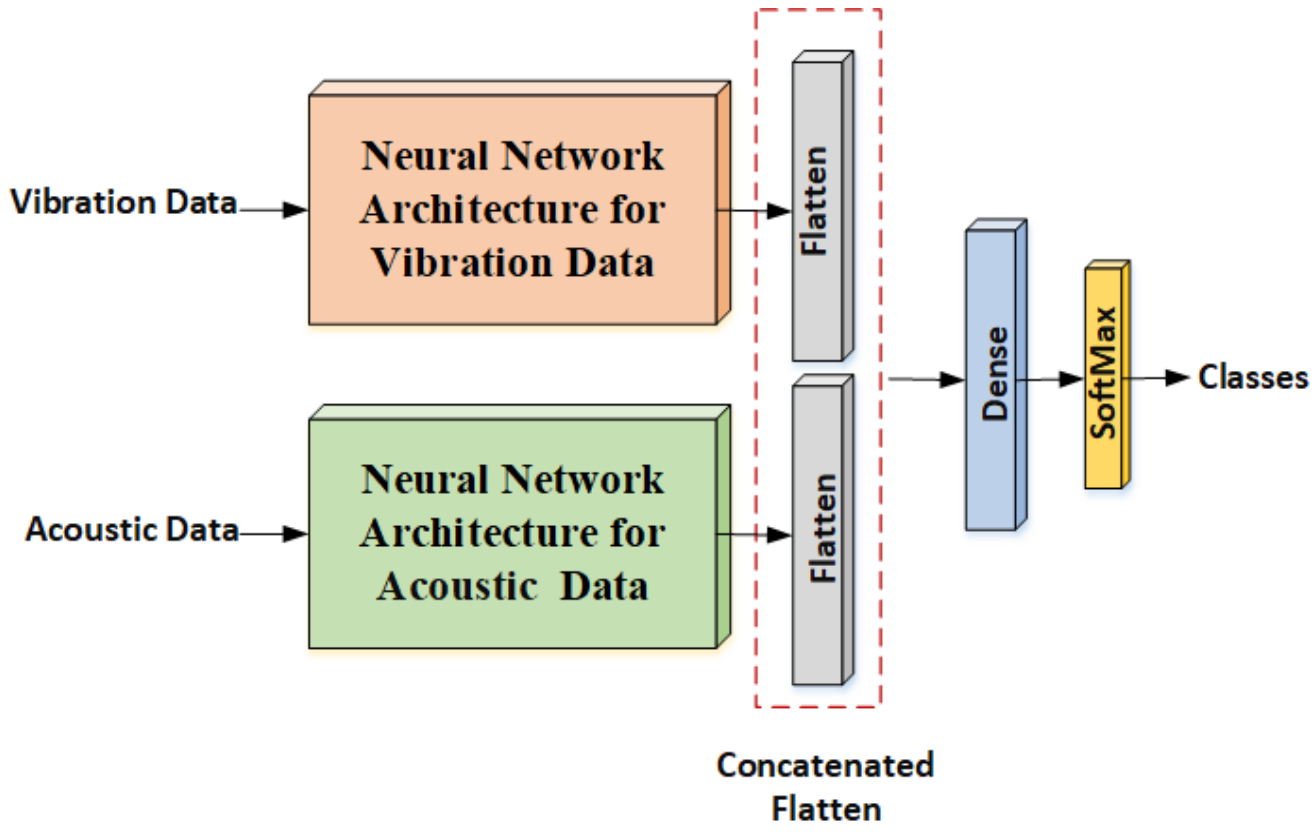

Figure 3. Neural Network Architecture for Combined Data

## III. Proposed Algorithm Results

The proposed algorithms in section 2 are executed using the rolling element bearing acoustic and vibration dataset presented in Table I. Figs. 4, 5, and 6 and Table II show the results obtained from the UORED-VAFCLS dataset, which consists of nine different classes with a 80% training and 20% validation split. This dataset consists of clean bearing signals that are healthy, fault-developing, and faulty. For vibration data alone, the maximum accuracy obtained was 93%. Looking at the confusion matrix, a majority of the classes are easily distinguishable except for the outer race fault developing and the outer race faulty data overlapping. For acoustics data results, shown in Fig. 5, a lower accuracy than the vibration data was achieved at 77%. When the confusion matrix was analyzed, similar results were obtained to those using the vibration data. Lastly, the proposed multi-modal acoustic and vibration data reached the highest accuracy at 95%.

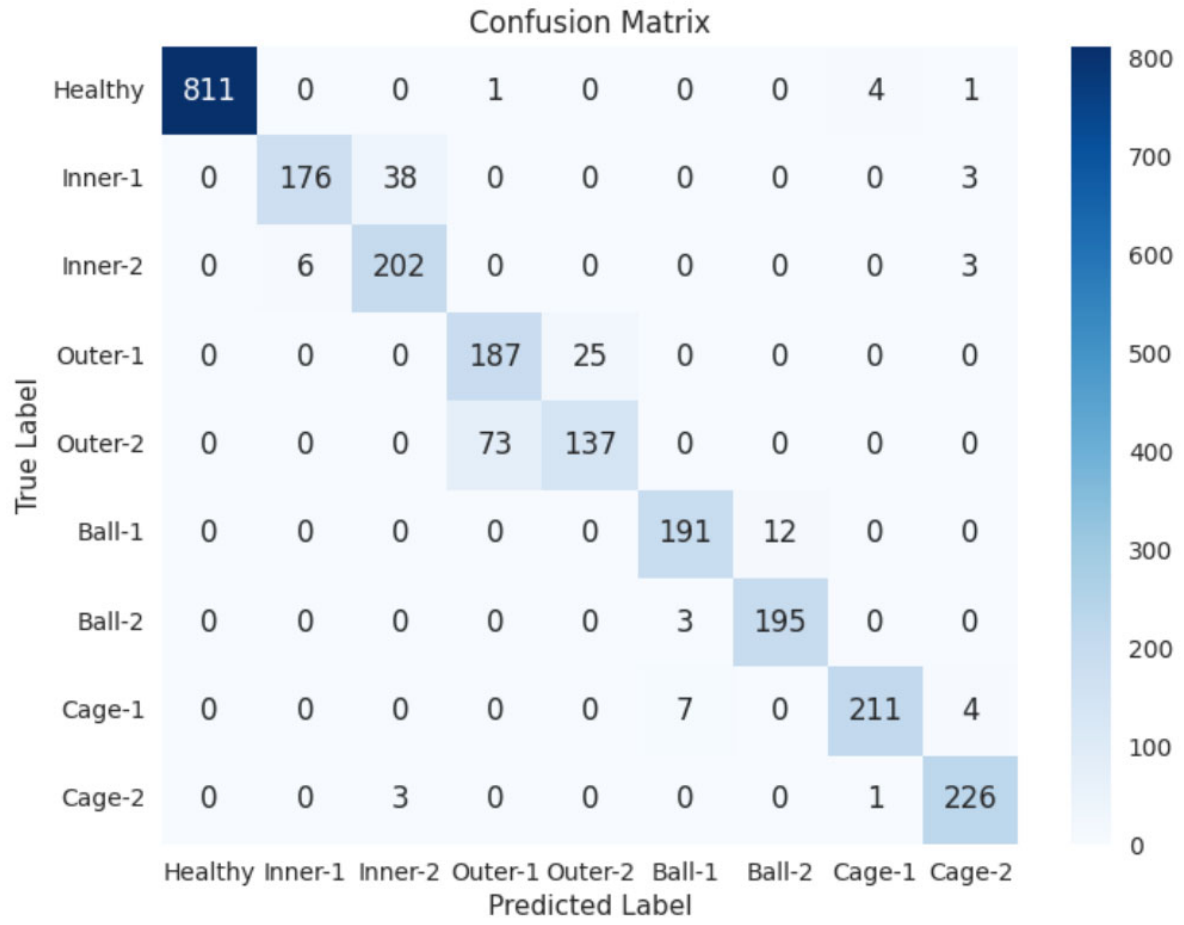

Figure 4. CNN Vibration Data Results for the UORED-VAFCLS Dataset

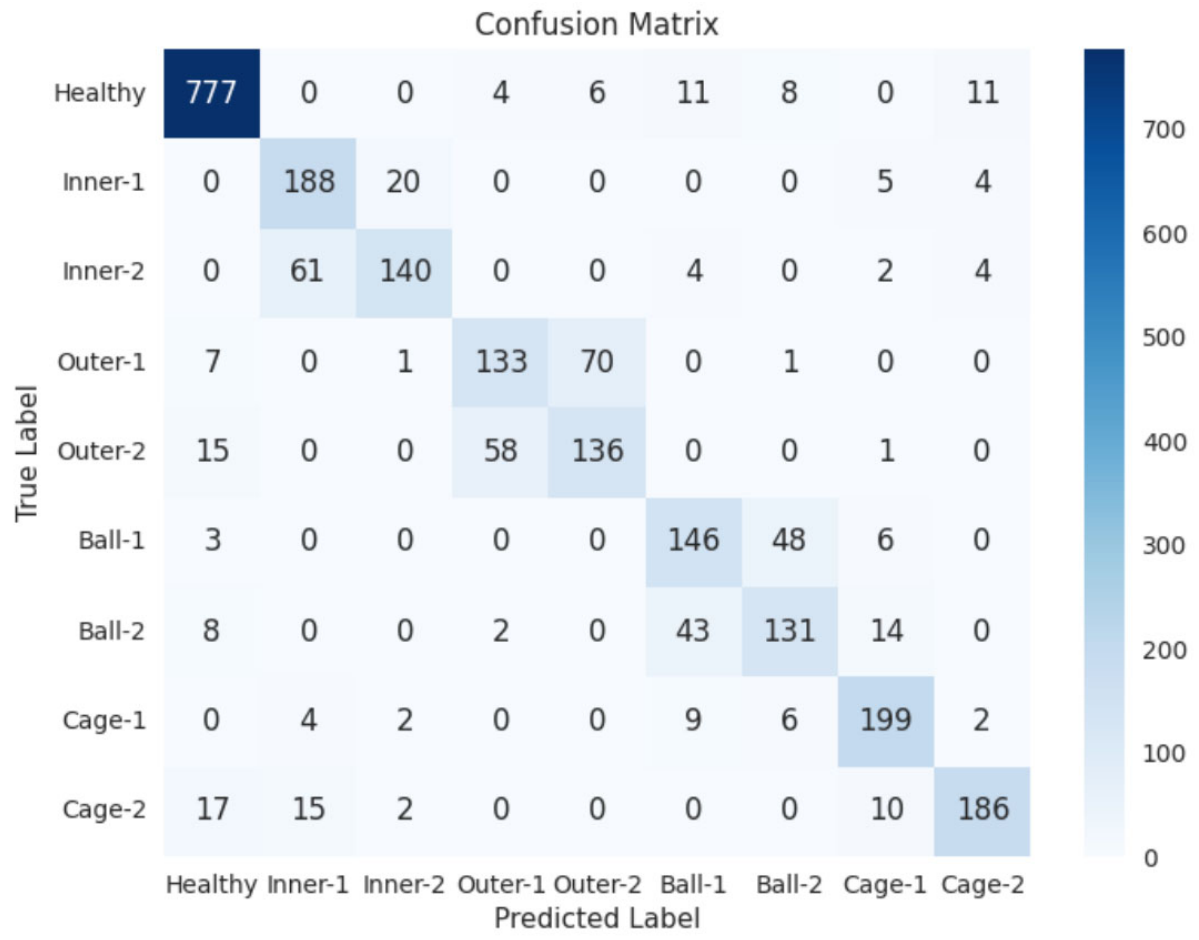

Figure 5. CNN and RNN Acoustic Data Results for the UORED-VAFCLS Dataset

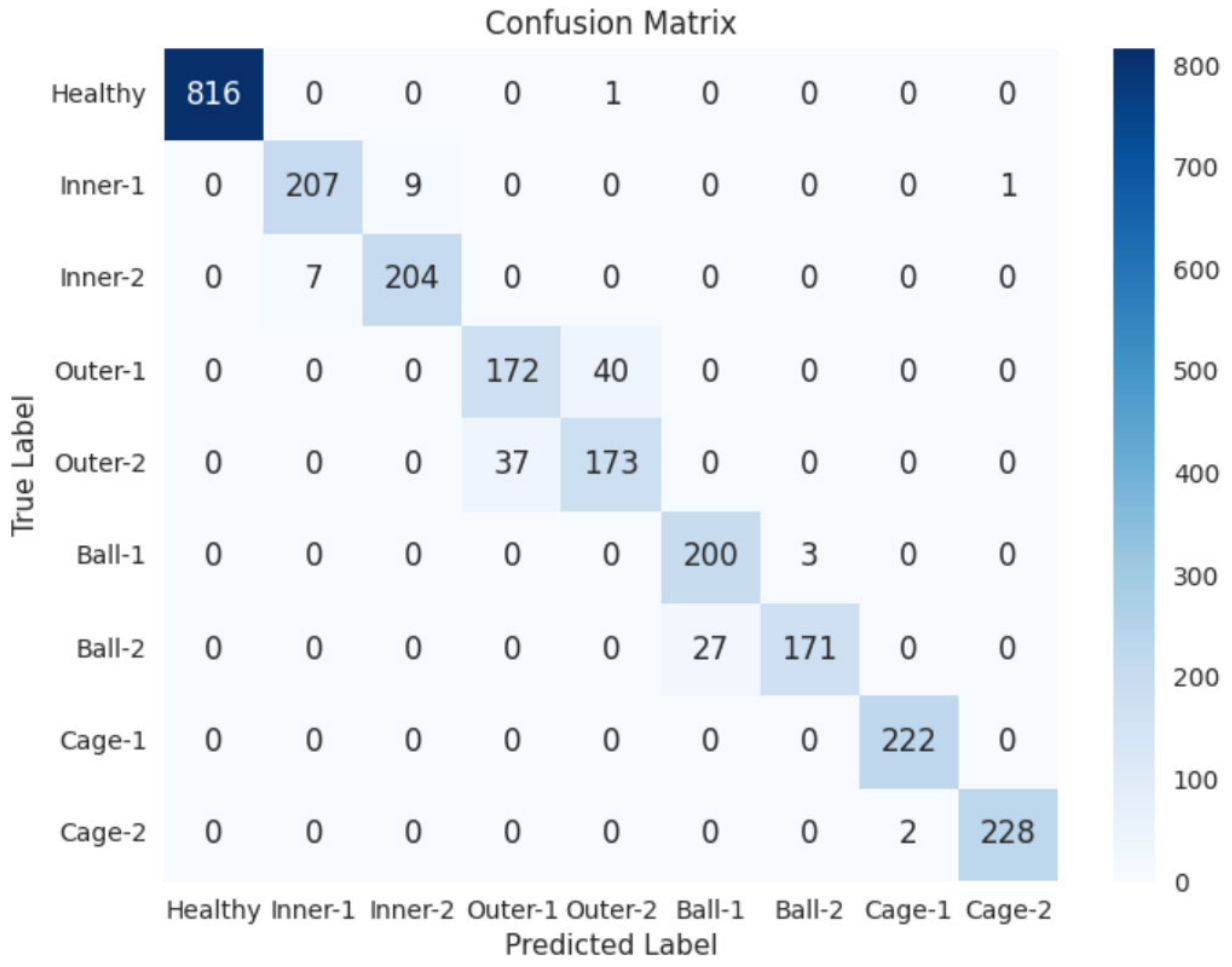

Figure 6. CNN and RNN Combined Acoustic and Vibration Data Results for the UORED-VAFCLS Dataset

TABLE II. UOERED-VAFCLS DATASET RESULTS

| | Classes | Precision (%) | Recall (%) | F1 (%) | Accuracy (%) |
|---|---|---|---|---|---|
| **Vibration** | Healthy | 100 | 99.26 | 99.63 | 99.76 |
| | Inner-1 | 96.70 | 81.10 | 88.22 | 98.13 |
| | Inner-2 | 83.12 | 95.73 | 88.98 | 98.01 |
| | Outer-1 | 71.64 | 88.20 | 79.06 | 96.07 |
| | Outer-2 | 84.56 | 65.23 | 73.65 | 96.11 |
| | Ball-1 | 95.02 | 94.08 | 94.55 | 99.12 |
| | Ball-2 | 94.20 | 98.48 | 96.29 | 99.40 |
| | Cage-1 | 97.68 | 95.04 | 96.34 | 99.36 |
| | Cage-2 | 95.35 | 98.26 | 96.78 | 99.40 |
| | Overall | 90.92 | 90.60 | 90.39 | 92.70 |
| **Acoustic** | Healthy | 93.95 | 95.10 | 94.52 | 96.42 |
| | Inner-1 | 70.14 | 86.63 | 77.52 | 95.67 |
| | Inner-2 | 84.84 | 66.35 | 74.46 | 96.19 |
| | Outer-1 | 67.51 | 62.73 | 65.03 | 94.32 |
| | Outer-2 | 64.15 | 64.76 | 64.45 | 94.04 |
| | Ball-1 | 68.54 | 71.92 | 70.19 | 95.07 |
| | Ball-2 | 67.52 | 66.16 | 66.83 | 94.84 |
| | Cage-1 | 83.96 | 89.63 | 86.71 | 97.57 |
| | Cage-2 | 89.85 | 80.86 | 85.12 | 97.42 |
| | Overall | 76.72 | 76.02 | 76.09 | 80.79 |
| **Vibration & Acoustic** | Healthy | 100 | 99.87 | 99.93 | 99.96 |
| | Inner-1 | 96.72 | 95.39 | 96.05 | 99.32 |
| | Inner-2 | 95.77 | 96.68 | 96.22 | 99.36 |
| | Outer-1 | 82.29 | 81.13 | 81.71 | 96.94 |
| | Outer-2 | 80.84 | 82.38 | 81.60 | 96.90 |
| | Ball-1 | 88.10 | 98.52 | 93.02 | 98.80 |
| | Ball-2 | 98.27 | 86.36 | 91.93 | 98.80 |
| | Cage-1 | 99.10 | 100 | 99.55 | 99.92 |
| | Cage-2 | 99.56 | 99.13 | 99.34 | 99.88 |
| | Overall | 93.41 | 93.27 | 93.26 | 94.96 |

The algorithms proposed in section II are then run using the electric motor acoustic and vibration dataset. Figs. 7, 8, and 9. Table III show the results obtained from the UORED-VAFCVS dataset, consisting of eight classes with a 80% training and 20% validation split. This dataset consists of clean induction motor signals that were healthy and seven induction motor faults. For vibration data alone, the maximum accuracy obtained was 99%; majority of the classes were classified using the confusion matrix. For acoustics data, shown in Fig. 8, results achieved a lower accuracy than the vibration data at 95%. When the confusion matrix is analyzed the results were similar to vibration data. Lastly, the proposed multi-modal combined acoustic and vibration data attained the highest accuracy at 100%. Although the results have a 1% difference in applications where a single sensor algorithm accuracy is low the proposed multi model can be applied to train two sensor inputs which will increase the algorithm accuracy.

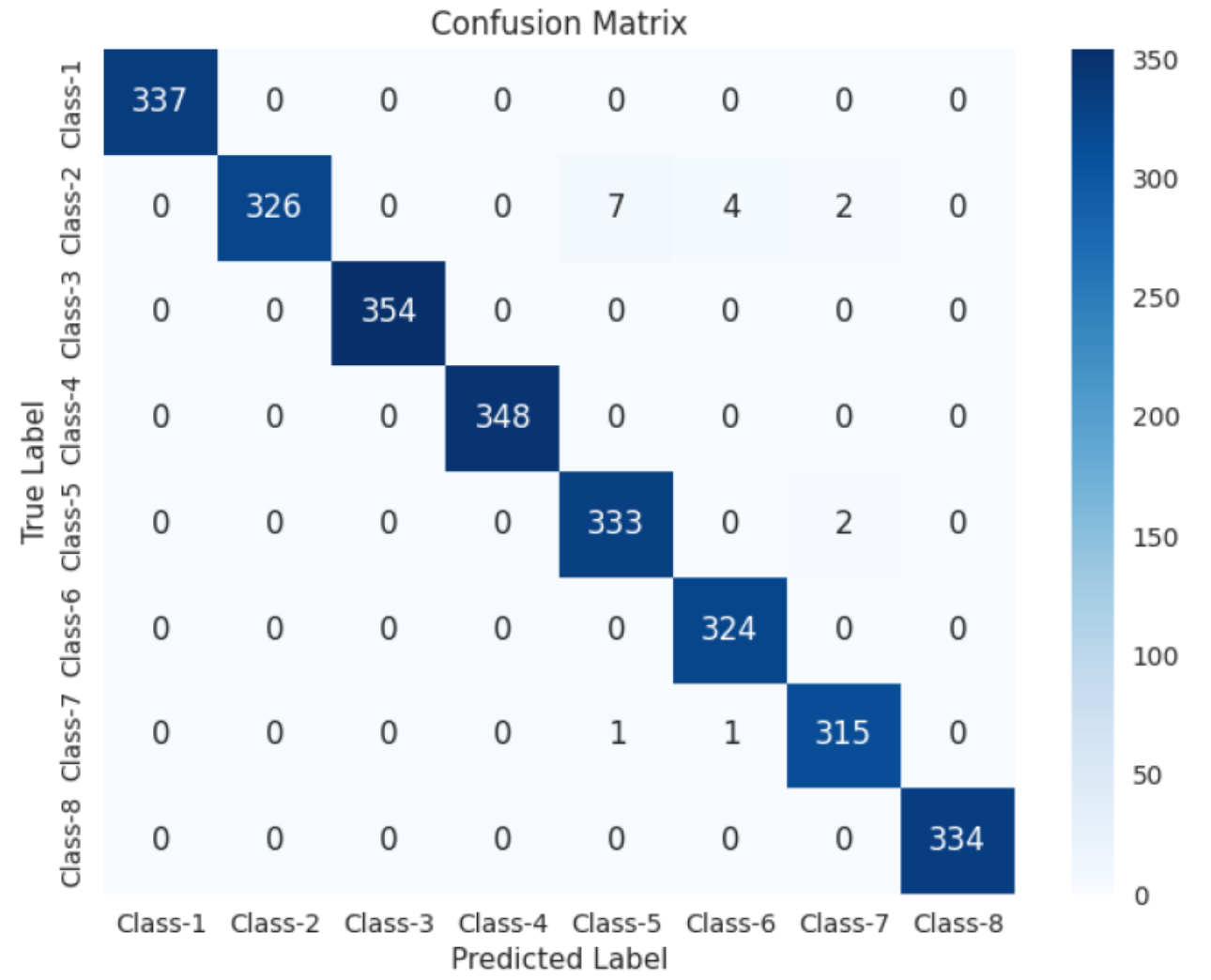

Figure 7. CNN Vibration Data Results for the UOEMD-VAFCVS Dataset

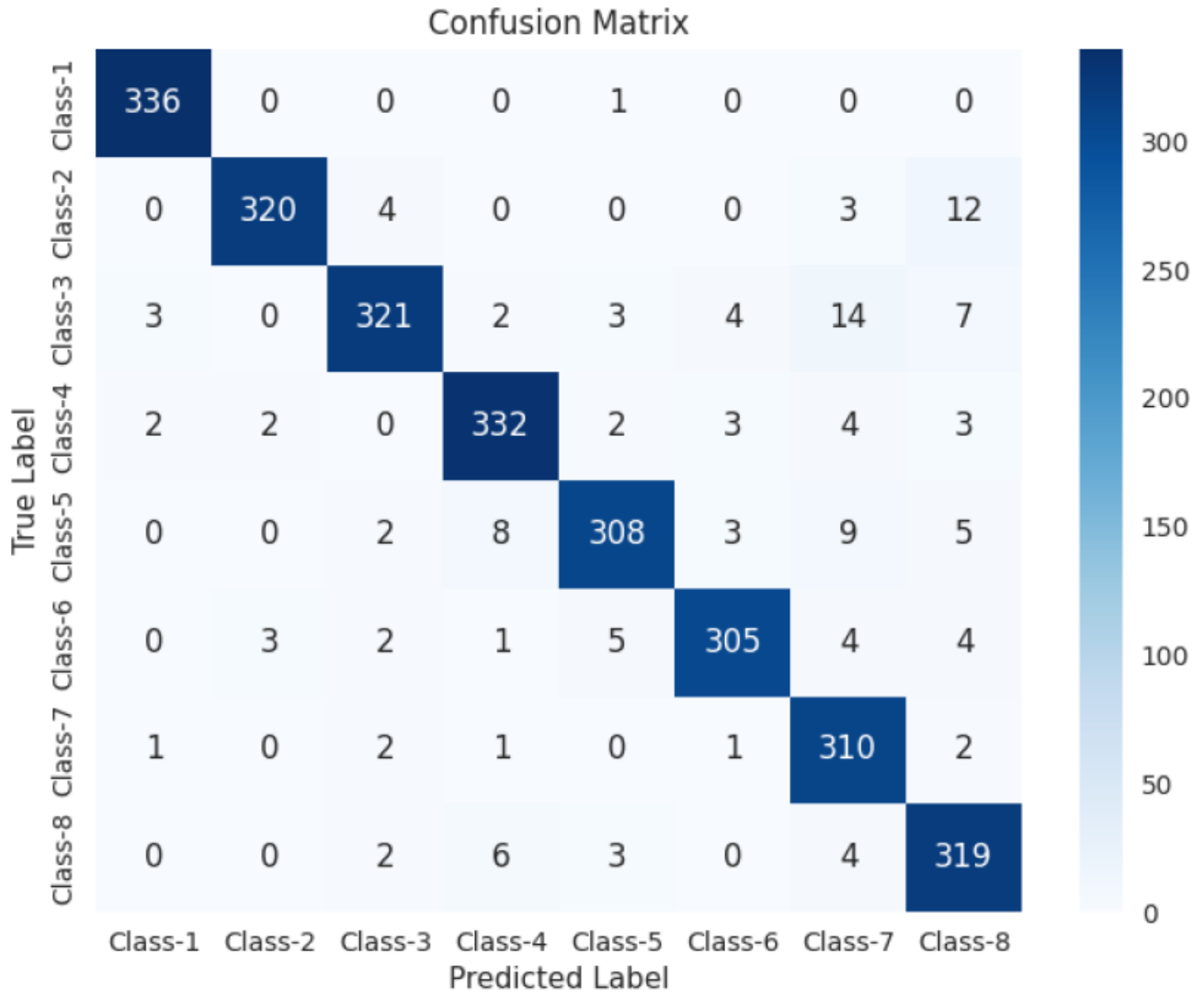

Figure 8. CNN and RNN Acoustic Data Results for the UOEMD-VAFCVS Dataset

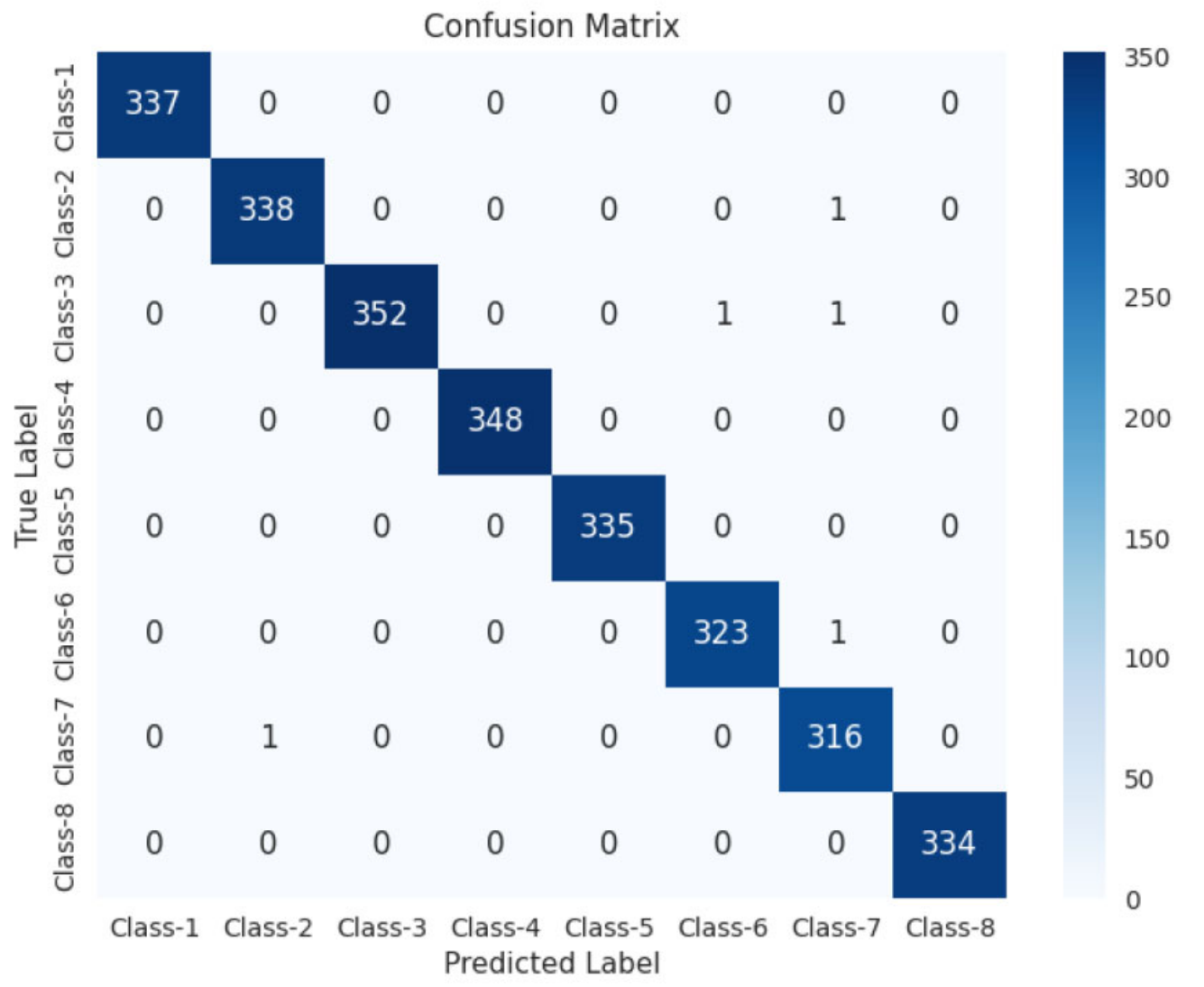

Figure 9. CNN and RNN Combined Acoustic and Vibration Data Results for the UOEMD-VAFCVS Dataset

TABLE III. UOEMD-VAFCVS DATASET RESULTS

| | Classes | Precision (%) | Recall (%) | F1 (%) | Accuracy (%) |
|---|---|---|---|---|---|
| **Vibration** | Class 1 | 100 | 100 | 100 | 100 |
| | Class 2 | 100 | 96.16 | 98.04 | 99.51 |
| | Class 3 | 100 | 100 | 100 | 100 |
| | Class 4 | 100 | 100 | 100 | 100 |
| | Class 5 | 97.65 | 99.40 | 98.52 | 99.62 |
| | Class 6 | 98.48 | 100 | 99.23 | 99.81 |
| | Class 7 | 98.74 | 99.36 | 99.05 | 99.77 |
| | Class 8 | 100 | 100 | 100 | 100 |
| | **Overall** | 99.36 | 99.36 | 99.35 | 99.37 |
| **Acoustic** | Class 1 | 98.24 | 99.70 | 98.96 | 99.73 |
| | Class 2 | 98.46 | 94.39 | 96.38 | 99.10 |
| | Class 3 | 96.39 | 90.67 | 93.44 | 98.32 |
| | Class 4 | 94.85 | 95.40 | 95.12 | 98.73 |
| | Class 5 | 95.65 | 91.94 | 93.75 | 98.47 |
| | Class 6 | 96.51 | 94.13 | 95.31 | 98.88 |
| | Class 7 | 89.08 | 97.79 | 93.23 | 98.32 |
| | Class 8 | 90.62 | 95.50 | 93.00 | 98.21 |
| | **Overall** | 94.97 | 94.94 | 94.90 | 94.90 |
| **Vibration & Acoustic** | Class 1 | 100 | 100 | 100 | 100 |
| | Class 2 | 99.70 | 99.70 | 99.70 | 99.92 |
| | Class 3 | 100 | 99.43 | 99.71 | 99.92 |
| | Class 4 | 100 | 100 | 100 | 100 |
| | Class 5 | 100 | 100 | 100 | 100 |
| | Class 6 | 99.69 | 99.69 | 99.69 | 99.92 |
| | Class 7 | 99.05 | 99.68 | 99.37 | 99.85 |
| | Class 8 | 100 | 100 | 100 | 100 |
| | **Overall** | 99.80 | 99.81 | 99.81 | 99.81 |

Integrating acoustic and vibration data into multi-modal analyses is a factor in enhancing the efficiency of machine learning algorithms. The UORED-VAFCVS dataset showcased that, while acoustic data alone may yield a lower accuracy at 95%, its integration with vibration data results in a remarkable accuracy of 100% for the detection of induction motor faults. This shows the importance of data fusion, demonstrating the significance of acoustic data in machine learning applications. The UOEMD-VAFCLS dataset further adds to this trend, where the results of acoustic data in a multi-modal approach enhanced the accuracy from 93% (vibration alone) to 95% for bearing fault detection. These findings shed light on acoustic data's essential role in enhancing the performance of machine-learning algorithms. Additionally, acoustic data from UOEMD-VAFCVS dataset had lots of noise when compared to UORED-VAFCLS dataset during data collection however the algorithm reached higher classification accuracy for the UORED-VAFCLS dataset which highlights that unprocessed noisy acoustic data can be used in noisy environments on top of vibration data to obtain better classification accuracies in machinery fault diagnosis.

## IV. Conclusion

In conclusion, this study highlights the significance of deep learning methodologies, particularly 1D convolutional neural networks (CNN) and long short-term memory (LSTM), in the realm of machine fault diagnosis. The study focused on using rolling element bearing and induction motor fault condition datasets. The proposed methodology uses vibration and acoustic data, specifically the UORED-VAFCVS and UORED-VAFCLS datasets, highlighting the importance of acoustic data in refining the accuracy of machine learning algorithms. The improvements in fault diagnosis accuracy through data fusion were achieved by combining vibration and acoustic data for the two datasets tested. As the machine learning field continues to evolve, the potential for deep learning using data fusion is advancing the precision of machine diagnosis, especially when multi-class datasets are taken into consideration.